\documentclass[runningheads]{llncs}
\usepackage[utf8]{inputenc}
\usepackage[T1]{fontenc}
\usepackage{multicol}
\usepackage[italian,american]{babel}
\usepackage{amsmath,mathtools,amssymb}
\usepackage[inline]{enumitem}
\usepackage{siunitx}
\usepackage{csquotes}
 \usepackage[ruled,vlined,linesnumbered]{algorithm2e}
\usepackage{subcaption}
\usepackage{tabularx}
\usepackage{booktabs,multirow}
\usepackage[
    left = \flqq{},% 
    right = \frqq{},% 
    leftsub = \flq{},% 
    rightsub = \frq{} %
]{dirtytalk}
\usepackage[sectionbib,numbers,sort&compress]{natbib}

\usepackage{graphicx}
\usepackage[hidelinks]{hyperref}
\usepackage{color}
\usepackage{url,cleveref}

\begin{document}
%%Insightful Mapping of Tree-Based Ensemble Models using Decision Predicate Graphs
\title{Decision Predicate Graphs: Enhancing Interpretability in Tree Ensembles}
\titlerunning{DPG for interpretability}
% Double-blind

\author{Leonardo Arrighi\inst{1}\orcidID{0009-0006-2494-0349} \and
Luca Pennella\inst{2}\orcidID{0009-0006-2721-1248}
\and 
Gabriel Marques Tavares\inst{3, 4}\orcidID{0000-0002-2601-8108}
\and
Sylvio Barbon Junior\inst{5}\orcidID{0000-0002-4988-0702}}

\authorrunning{L.~Arrighi et al.}
\institute{
Department of Mathematics and Geosciences, University of Trieste, Trieste, Italy\\
\email{leonardo.arrighi@phd.units.it} \and
Department of Economics, Business, Mathematics and Statistics, University of Trieste, Italy \\
\email{luca.pennella@phd.units.it} \and
LMU Munich, Munich, Germany \and
Munich Center for Machine Learning (MCML), Munich, Germany \\
\email{tavares@dbs.ifi.lmu.de} \and
Department of Engineering and Architecture, University of Trieste, Trieste, Italy \\
\email{sylvio.barbonjunior@units.it}}
\maketitle              % typeset the header of the contribution
\begin{abstract}
Understanding the decisions of tree-based ensembles and their relationships is pivotal for machine learning model interpretation. Recent attempts to mitigate the human-in-the-loop interpretation challenge have explored the extraction of the decision structure underlying the model taking advantage of graph simplification and path emphasis. However, while these efforts enhance the visualisation experience, they may either result in a visually complex representation or compromise the interpretability of the original ensemble model. 
In addressing this challenge, especially in complex scenarios, we introduce the Decision Predicate Graph (DPG) as a model-agnostic tool to provide a global interpretation of the model. DPG is a graph structure that captures the tree-based ensemble model and learned dataset details, preserving the relations among features, logical decisions, and predictions towards emphasising insightful points. Leveraging well-known graph theory concepts, such as the notions of centrality and community, DPG offers additional quantitative insights into the model, complementing visualisation techniques, expanding the problem space descriptions, and offering diverse possibilities for extensions. Empirical experiments demonstrate the potential of DPG in addressing traditional benchmarks and complex classification scenarios.

\end{abstract}
\newpage
\section{Introduction}\label{Intro}
Artificial intelligence, although still under strong development, is now a consolidated and widely used tool. 
This is thanks to the continuous growth of computing power, which allows the use of increasingly complex and computationally expensive machine learning (ML) methods. 
%Additionally, the diffusion of technological tools and sensors enables the collection of large quantities of data.
The challenges presented by modern-world problems are growing in complexity as well as the proposed solutions.

Dealing with large quantities of data and frequently encountering unbalanced datasets are still significant obstacles in addressing many real-world issues; however, tree-based ensemble algorithms offer several advantages in overcoming these challenges, including robustness to noise and outliers, scalability to large datasets, automatic handling of missing values, and the ability to capture complex relationships and interactions within the data \cite{malekloo_machine_2022, mienye_survey_2022}.
%These models have versatile applications spanning diverse scenarios, ranging from soil science \cite{chen_comparative_2017} and computational biology \cite{mohammadi_ensemble_2022} to construction \cite{li_high-performance_2022}, sustainable computing \cite{p_prediction_2022}, agricultural \cite{pereira2018predicting,viana_evaluation_2021}, and food security \cite{barbon2016storage,dharmawan_study_2022}.

Without delving into the intricacies of various algorithms, exhaustively described by \citet{hastie_additive_2009}, the process of learning tree-based ensemble models involves training multiple decision trees to combine their predictions, optimising performance, and enhancing generalisation capabilities. 
While these models indeed offer concrete solutions to a diverse array of problems, developers and users are confronted with new challenges. 
The training outcome yields an exceptionally complex model, commonly referred to as an opaque-box model (also known as the black-box problem), whose internal workings are not transparent or easily interpretable \cite{chimatapu2018explainable}.
In response to this context, the Decision Predicate Graph (DPG) is proposed in this paper.

Drawing inspiration from the expanding theme of eXplainable Artificial Intelligence (XAI), we designed a graph structure to tackle transparency and explainability challenges inherent in tree-based ensemble models. 
This facilitates a better understanding of the intricate choices underlying these ML models.
DPG is created with inspiration from the concept of aggregating random forests (RF) \cite{ho_random_1995, breiman_random_2001}, as introduced by \citet{Gossen2023267}. 
The approach proposed in \cite{Gossen2023267} suggests visualising the decisions within the RF by combining the branches of the tree base learners into a single and compact decision diagram.
The concept behind DPG is to convert a generic tree-based ensemble model for classification into a graph, a defined and studied structure with known properties. 
In this graph, nodes represent predicates, i.e., the feature-value associations present in each node of every tree, while edges denote the frequency with which these predicates are satisfied during the model training phase by the samples of the dataset.
The DPG structure enables comprehending the choices made by the model, enhancing transparency and understandability.
Moreover, it allows the exploitation of graph properties to develop metrics and algorithms facilitating the analysis of the ensemble model. 
This, in turn, aids in understanding the decisions it makes, easing the task of visualising the graph which can be vast and complicated for larger models with numerous tree base learners.

DPG serves as a model-agnostic tool offering a comprehensive interpretation of tree-based ensemble models. 
It provides descriptive metrics that enhance the understanding of the decisions inherent in the model, offering valuable insights. 
This tool proves particularly useful for models that are \emph{a priori} considered satisfactory in terms of performance.
% In this sense, DPG serves as an analysis tool applied to the trained model. By employing DPGs through the mapping of decision trees into a graph structure, we can attain an interpretation of the algorithm's operation. 
% This approach facilitates the identification of the most relevant features and enables the extraction of the crucial 'trace' in terms of paths executed by the sample.
% Further, DPGs precisely reflect the semantics of the original ensemble method and consider many related methods as working together rather than competing because they produce different types of results. 
% For example, combining SHAP values, LIME or Anchors with our approach can create a more complete explanation and have their information complemented by DPG metrics. 
% Its utility lies in the fact that it doesn't impact the training of the model, providing valuable insights into a model known \emph{a priori} to be satisfactory in terms of performance.

Our work contributes in the following ways:
\begin{itemize}
    \item we introduce DPG, a novel interpretability structure that transforms an opaque-box tree-based ensemble model into an enriched graph;
    \item we present the algorithm used to create DPG, accompanied by pseudo-code to enhance understanding and facilitate replication, complete with its asymptotic complexity;
    %\item we provide various graph analysis tools to comprehend the decisions made by the model;
    \item we provide the interpretation of three metrics from graph theory, enriching the  model comprehension and gaining insights;
    \item we demonstrate the use of the proposed method through two case studies: the application of DPG to two RF models, respectively, on the Iris dataset \cite{fisher1936use} and a challenging dataset.
    %\item we discuss the current limitations of this tool and propose potential future developments.
\end{itemize}

It is important to highlight that these results are achieved in a generic fashion, utilising a standard classifier on well-established datasets.
Significantly, we intentionally avoided incorporating scenario-specific heuristics. Therefore, we posit that our aggregation approach has the potential for widespread application across a diverse range of related scenarios.

\section{Literature Review}\label{LR}

%In contemporary machine learning, models find extensive application across diverse domains, demonstrating commendable performance. 
Complex yet effective models have become increasingly prevalent, especially in fields where the outcomes bear significant importance and require a heightened level of sensitivity.
% as mentioned in Section~\ref{Intro}. 
In these scenarios, there is a growing demand for models to exhibit transparency, accountability, and a comprehensive understanding. 
Consequently, the discussion on XAI has experienced significant growth, leading to an extensive body of literature \cite{guidotti2018survey,hanif2021survey,adadi2018peeking}.
%such as economics \cite{Nosratabadi20201,Popat20181120,Amarasinghe2023} and healthcare \cite{Jiang2017230,Cutillo2020,vstrumbelj2010explanation}.
%According to \citet{Dwivedi}, a widely acknowledged criterion for categorising XAI pertains to their scope: global explanations, aimed at elucidating overarching trends in the data and providing an understanding of the entire model; and local explanations, which seek to illuminate the rationale behind a specific prediction for a given instance. Another criterion involves discerning whether interpretability is achieved by constraining the complexity of the machine learning model (\emph{intrinsic}) or by implementing methods that scrutinize the model post-training (\emph{post hoc}).

%Intrinsic interpretability refers to machine learning models deemed interpretable due to their straightforward structures, such as concise decision trees or linear models. On the other hand, post hoc interpretability involves the application of interpretation methods subsequent to model training \cite{adadi2018peeking}. Additionally, methods for interpretation can be categorised as model-specific or model-agnostic.
In XAI, as suggested by \citet{Dwivedi}, classifications are often based on scope, distinguishing between global interpretability, which reveals overall data trends and provides insights into the entire model, and local interpretability, which elucidates the reasoning behind specific predictions for individual instances. 
Another classification criterion involves how interpretability is achieved: intrinsic interpretability relies on straightforward model structures (e.g., concise decision trees or linear models), while post hoc interpretability involves methods applied after model training \cite{adadi2018peeking}.
Interpretation methods are further categorised as model-specific or model-agnostic.
Model-specific interpretation tools are tailored to specific model classes, limiting their applicability.
In contrast, model-agnostic tools are versatile, capable of being employed with any ML model, and are applied post-training (e.g., SHAP \cite{Lundberg20174766}, LIME \cite{ribeiro2016should}, PDP \cite{friedman2001greedy}, and Anchors \cite{ribeiro_anchors_2018}).

In this context, XAI tools, especially those providing global interpretations, become valuable instruments for understanding tree-based ensemble models.
These models are widely used in addressing diverse problem domains, as highlighted in several surveys \cite{chipman1998making,Haddouchi2019,aria2021comparison}.
% Numerous studies have delved into model-specific techniques tailored for ensembles of trees.  Below, we describe the contribution of these papers, and then we highlight the distinctive features of our work compared to those of these studies. 
As a result, the number of studies delving into model-specific techniques tailored for ensembles of trees has also increased.

The first significant study is proposed by \citet{Mashayekhi2015223}. 
The authors introduced \emph{RF+HC}, an approach that employs a hill climbing algorithm in RF to search for a decision set.
This rule set reduces the number of decisions dramatically, which significantly improves the comprehensibility of the underlying model built by RF.
Similarly, \citet{hara2018making} exploits Bayesian model selection to extract the decision set.
These approaches share similarities with our method; however, our proposal extends beyond the extraction of decisions from the RF.
Visualising the decisions of tree-based ensemble models and simultaneously complementing them with metrics developed by graph theory makes DPG more adaptable and holistic. This approach allows for obtaining insights into the model beyond just decision information.
% Using various metrics, we provide additional insights 
% We provide additional insights into the algorithm's structure, predicates, and crucial paths using various metrics. 
% Moreover, for smaller RF, our method allows convenient visualisation.

\citet{Zhao2019407} proposed \emph{iForest}, a visual analysis system specifically designed to interpret RF models and their predictions. 
They built a feature view to illustrate the relationships between input features and outcome predictions and proposed a design that summarises multiple decision paths based on feature occurrences and ranges, allowing users to explore and understand the partitioning logic of these paths.
The iForest, like numerous visualisation systems, faces significant challenges related to scalability and interpretability when dealing with large ensembles.
To overcome this challenge, our approach does not prioritise visualisation but instead strives for a comprehensive understanding of the ensemble's logic. 
%This includes metrics that provide valuable insights even in the presence of complex tree-based ensemble models.

\citet{Hatwell20205747} contributed with \emph{Collection of High Importance Random Path Snippets} (CHIRPS), a method that incorporates the explanation of RF classification for each data instance and extracts a decision path from each tree in the forest, resulting in a set of decisions that elucidate the classification process.
However, this method is limited to rules extraction and lacks insight into the model's structure.
Additionally, it does not incorporate metrics to explain the logic of the tree-based ensemble model.

Another technique focused on visualising decisions underlying RF models is introduced by \citet{neto2020explainable}. 
\emph{Explainable Matrix} (ExMatrix) employs a matrix-like visual metaphor, where rows represent decisions, columns denote features, and cells encapsulate decision predicates, thereby facilitating the scrutiny of models and the audit of classification outcomes. 
The visualisation capability of ExMatrix for global visualisation has limitations in terms of scalability because the number of decisions increases significantly with the number of trees in large ensembles.
Moreover, ExMatrix layouts can rapidly become challenging to explore, while the complexity of the model increases. 
As mentioned earlier, our approach enables us to retrieve information without relying solely on visualisation.

\citet{Dedja20231} introduced an approach denoted as \emph{Building Explanations through a LocalLy AccuraTe Rule EXtractor} (BELLATREX), which is designed to explain the forest predictions for a given test instance by a set of logical rules based on the features of the dataset.
%The BELLATREX method aims to explain the predictions of an RF using a few rules.
However, a potential limitation lies in the computational complexity of the approach. 
While explaining a single prediction is quick, applying the method to a complete dataset becomes computationally expensive.
Furthermore, BELLATREX focuses on instance-level explanations rather than providing a global perspective.
In addition, BELLATREX uses clustering techniques to simplify the representation and decision logic, whereas our approach uses graphs to avoid simplifications that can lead to loss of information.

Various studies \cite{van2007seeing,zhou2016interpreting,deng2019interpreting,gulowaty_extracting_2021} proposed several tree similarity metrics through the process of clustering based on tree representations.
However, these approaches, while beneficial for interpretability, require simplifying the model through techniques such as selection, pruning, and frequency analysis, which can result in information loss.

A number of works \cite{Nakahara2017266,silva_rdsf_2023, Gossen2023267,murtovi_forest_2023} established a connection between tree-based ensemble models and graph theory.
The foundational concept of these techniques has been explored by \cite{oliver1992decision,8372043,pmlr-vR3-needham01a,Tan2002131,Florio20237577,Zhu202113707,wallace2005statistical}. These works established the theory that introduces the transformation of decision trees into graphs, aiming for more efficient and non-redundant tree structures.
\citet{Nakahara2017266} and \citet{silva_rdsf_2023} works focus on performance optimisation, with \citet{Gossen2023267} and \citet{murtovi_forest_2023} being the sole contributors that employed these techniques for interpretability purposes.

In particular, \citet{Gossen2023267} introduced the \emph{Algebraic Decision Diagram} (ADD), aiming to transform tree-based ensemble models into bipartite graphs.
The ADD serves as an alternative construction to RF, providing an additional predictive model that functions as a surrogate.
ADD proves valuable for specific aspects of interpretability, such as outcome explanation, logic of \emph{majority vote}, and visualising the path. 
Acting as a surrogate model, their primary contributions lie in the ability to provide a simple, optimised model.
In comparison, our objective is to extend their ideas by fully leveraging graph theory.
This goes beyond the transposition of tree-based ensemble models into a graph; rather, it involves using graph theory and its associated metrics to gain insights into the functionality of the models.

\emph{ForestGUMP}, an online tool developed by \citet{murtovi_forest_2023}, is designed for generating ADDs. 
This tool provides valuable information such as graph visualisation, hypothetical sample path display, and logic of majority vote.
However, it has some limitations in terms of the number of usable trees (only \num{20}) and in visualising problems that involve multiple features and classification choices, making graph navigation complex.
In our work, while we enable visualisation, our focus is on utilising graph-related metrics, and we do not incorporate the simplification of the analysed models.

%In comparison to the aforementioned papers, our objective is to extend their ideas by fully leveraging graph theory, transcending the limitation of bipartite graphs. 
%While bipartite graphs enhance visibility and simplify the problem, they fall short in making the most of metrics associated with graph theory, such as communities, degree, eigenvector centrality, and LRC. 
%Our aim goes beyond the transposition of tree-based ensemble models into a graph; rather, it involves using graph theory and its associated metrics to gain insights into the functionality of the models.

\section{Decision Predicate Graphs}\label{sec:DPG}

DPGs capture the details of a tree-based ensemble model and learned dataset specifics, emphasising predicate paths while preserving crucial decision points. 
Essentially, DPG converts complex ensemble models into a graph structure, where nodes represent predicates made by the model and edges denote the occurrence of these predicates during model training. 

In this segment, we present the formalisation of DPG, elucidate the algorithm employed in its development through pseudo-code and its asymptotic complexity, and provide a detailed exposition of various metrics and properties essential for comprehending the tree-based ensemble model.
Moreover, we meticulously outline the advantages of metrics and articulate why they can serve as a valuable complementary aid to graph visualisation, particularly in overcoming its inherent limitations.

As previously mentioned in \Cref{Intro}, DPG is tailored for tree-based ensemble models designed specifically for classification tasks.

\subsection{Definition}
Let $\mathcal{M}_n$ be the tree-based ensemble model consisting of $n$ tree base learners $T(x; \Theta_b)$, where $x$ is a generic sample, and $\Theta_b$ characterizes the $b$th learner in terms of split variables, cutpoints at each node, and terminal-node values.
More specifically, $\Theta_b$ includes:
\begin{itemize}
    \item all the splitting conditions associated with each $j$th internal node $n_{bj}$ based on a specific feature $f_{bj}$ and a threshold (for numerical features) or a set of possible values (for categorical features) $t_{bj}$;
    \item the values assigned to leaf nodes $c_{b}$.
\end{itemize}
Let $\mathcal{D}$ be the training set on which $\mathcal{M}_n$ is trained.
Every base learner is trained on a dataset $\mathcal{D}_b$, where $\mathcal{D}_b$ is a subset of $\mathcal{D}$. We define $\mathcal{O}$ as the set of logical operations: $\mathcal{O}=\{\leq, >, =, \neq \}$.

A predicate set $\mathcal{P}(\mathcal{M}_n)$, for an ensemble method $\mathcal{M}_n$ is the set obtained by the union of the set of all the triples $p=(f_{bj}, o, t_{bj})$, where $o\in\mathcal{O}$, and $f_{bj}$, $t_{bj}\in\Theta_b$, and the set of all leaf nodes $c_b$, for all $n$ tree base learners of $\mathcal{M}_n$. The triples $p$ are called decisions, while the elements of $\mathcal{P}(\mathcal{M}_n)$ are called predicates. 

A Decision Predicate Graph (DPG$(\mathcal{M}_n)$) for a model $\mathcal{M}_n$ is a directed weighted graph $(\mathcal{P}, E)$ where:
\begin{itemize}
    \item $\mathcal{P}$ is the set of nodes, which corresponds to the predicate set $\mathcal{P}(\mathcal{M}_n)$;
    \item $E$ is the set of edges, where each edge represents the frequency with which a sample consecutively satisfies two predicates in a given base learner. This frequency is computed considering all the samples of the training set $\mathcal{D}$ for each base learner. 
    
\end{itemize}
For conciseness, from this point onward, we will use the acronym DPG to refer to the graph, indicating that it is constructed based on a model.

\subsection{From Ensemble to a DPG}\label{sec:pseudo_code}

We introduce an algorithm, outlined in Algorithm \ref{pseudocode}, for constructing the DPG based on a tree-based ensemble model, by traversing all tree base learners with the training samples.

\begin{algorithm}[H]
\SetKwFunction{traversing}{TRAVERSING}
\SetKwFunction{aggregating}{AGGREGATING}
\label{pseudocode}
\caption{Construct DPG from Ensemble Tree Model}

\KwIn{Ensemble tree model $\mathcal{M}_n$, Training set $\mathcal{D}$}
\KwOut{DPG$(\mathcal{M}_n)$}

Initialise empty set DPG$(\mathcal{M}_n)$\;

\ForEach{$T$ (learner) \textbf{in} $\mathcal{M}_n$}{
    
Initialise empty predicate set $\mathcal{P}$ and edge set ${E}$\;
      
    \ForEach{$x$ (sample) \textbf{in} $\mathcal{D}$}
    {
        Initialise empty predicate set $\mathcal{P}_x$ and edge set ${E_x}$\;
        \tcp{To obtain the predicates path for $x$ on the tree $T$}
        ($\mathcal{P}_{x}$, ${E_x}$) $\leftarrow$ \traversing{$T, x$}\;
        Add ($\mathcal{P}_{x}$, ${E_x}$) to $\mathcal{P}$ and ${E}$\;
    }
}

DPG$(\mathcal{M}_n)$ $\leftarrow$ \aggregating{($\mathcal{P}$, ${E}$)}\;

\Return{DPG$(\mathcal{M}_n)$}\;
\end{algorithm}

The algorithm iterates over each base learner in the ensemble tree model $\mathcal{M}_n$ and each training sample $x$ in the training set $\mathcal{D}$. %For each sample, the algorithm traverses the decision tree $\theta$ using the function \texttt{TRAVERSING}, which identifies the sequence of predicates representing the decision path taken by the sample. These predicates are aggregated across all samples to construct the DPG$(\mathcal{M}_n)$ using the \texttt{AGGREGATING} function.

To clarify, the \texttt{TRAVERSING} function follows the predicate path of a particular input sample $x$ through the decision tree $T$, starting from the root node and navigating to the appropriate leaf node based on the feature values of $x$. Meanwhile, the \texttt{AGGREGATING} function processes the predicates and edges obtained from \texttt{TRAVERSING} the decision trees for all samples into a single graph representation, DPG$(\mathcal{M}_n)$, by taking the union of $\mathcal{P}$ and computing the frequency of elements of $E$.

The algorithm presents a systematic methodology for constructing DPG.
% , thereby enhancing interpretability and offering deeper insights into the decision-making processes of these models. 
The overall asymptotic complexity can be formally expressed as follows:

\[
O(b \times s \times (k + k^2)) = O(b \times s \times k^2)
\]

where:
\begin{itemize}
    \item \( b \) is the number of learners in the ensemble,
    \item \( s \) is the number of samples in the training set, $|\mathcal{D}|$, and
    \item \( k \) represents the size of the  ($\mathcal{P}_{x}$, ${E_x}$) processed by the \texttt{TRAVERSING} and  \texttt{AGGREGATING} functions.
\end{itemize}

This analysis takes into account the linear time complexity \(O(k)\) for the \texttt{TRAVERSING} function and the quadratic time complexity \(O(k^2)\) for the  \texttt{AGGREGATING} function. %Our \emph{Python} 3.10 implementation, incorporating essential libraries such as \emph{graphviz} for visualisation, \emph{itertools} for seamless iteration over combinations, \emph{networkx} for graphical representation, and \emph{collections} for precise counting of occurrences, is accessible here\footnote{https://github.com/sbarbonjr/fhg/tree/main}.
Our \emph{Python} 3.10 implementation is accessible here\footnote{https://github.com/sbarbonjr/fhg/tree/main}.

\subsection{DPG interpretability}\label{DPG_inter}
In this section, we enumerate and elucidate some of the advantages that DPG can offer. 
The graph-based nature of DPG provides significant enhancements in the direction of a complete mapping of the ensemble structure. 
Weighted directed graphs, such as DPGs, are studied structures with well-established properties that enable the identification or construction of useful metrics and algorithms. 
It is crucial to emphasise that all the observations presented in this section are valuable for comprehending and analysing the obtained model.

\subsubsection{Visualisation.}\label{sec:viz}
DPG provides an immediate advantage by allowing the visualisation of the entire tree-based ensemble models through a single comprehensive graph.
Similar to the idea proposed by \citet{Gossen2023267}, consolidating all individual basic learners within the model into a unified graph provides a holistic representation of the decision-making process. 
This visualisation not only elucidates the decisions made by the learners but also reveals the intricate relationships between them. 
Consequently, it facilitates a comprehensive understanding of the utilised features and, more importantly, the associations between features and their values that enable the model to accurately classify a sample into a specific class.

Another noteworthy aspect of DPG lies in the concept of representing edges as frequencies.
This feature enables a discerning analysis of the most significant path through predicates, shedding light on decisions consistently employed by numerous learners or across multiple samples. 
This insight not only highlights the prevalence of certain decisions but also opens avenues for targeted enhancement strategies, focusing on those influential aspects within the model.

Moreover, by traversing all the possible paths between predicates in reverse, starting from one of the classes, we can discern the essential characteristics that a sample must possess to be classified into a specific class.
This capability facilitates the \emph{a priori} elimination of certain elements from the dataset when considering the particular class.

Nevertheless, we acknowledge that while visualisation is a valuable tool, its effectiveness diminishes with an increasing number of tree base learners. 
A multitude of tree base learners implies an increase in decisions and, consequently, an abundance of predicates. 
As a result, the size of the graph, in terms of nodes, grows proportionally with the model's scale.
This expansion can render the graph illegible or impractical to visualise due to its intricate complexity.
To address this challenge, we provide additional tools that complement the visualisation, aiding in the extraction of model properties and facilitating a more comprehensible understanding.

One approach is based on the desire and feasibility of determining the specific characteristics a sample must exhibit to be assigned to a particular class. 
Taking inspiration from the \emph{outcome explanation problem} introduced by \citet{Gossen2023267}, to enhance the immediacy and effectiveness of this analysis, we provide an aggregation of predicates, referred to as \emph{constraint}, which represent intervals associated with the features of each class.
The constraints are defined as follows: for a given class identified in the DPG, we list all nodes connected by a path originating from the node itself and culminating in the class. 
For each feature within the node predicates, we delineate the most extensive possible interval using the values associated with the features. 
This interval is defined by two endpoints. 
The minor endpoint is the smallest value within the set of values less than the feature, while the major endpoint is the largest value within the set of values greater than the feature. 
If either of these two sets is empty, the interval is deemed infinite.
Each class has its constraints for every feature contributing to the classification of the samples.

% Additional tools will be described in the following subsections.

% \subsubsection{Cut-Vertices}\label{cut}
% According to \citet{al-taie_python_2019}:
% \begin{displayquote}
%     A cut-vertex is a vertex that if removed, the number of network components increases,
% \end{displayquote}
% where a component is a weakly component of the directed graph, that is, a maximal subgraph in which there exists a directed path between any pair of vertices, but the direction of the edges is ignored. 
% Upon observation, if we identify a cut-vertex within the DPG and both connected components resulting from the removal of the edge contain at least one leaf node, it means that a predicate (i.e., the cut-vertex) can entirely separate the two or more classes represented by the leaf nodes. In this scenario, we can posit that the entire ensemble method is characterised by a collective decision made by all tree base learners, effectively dividing the mentioned classes. Recognising this information proves to be considerably useful as it allows us to reduce the complexity of the problem. Moreover, it's important to note that the cut-vertex is not necessarily a root.

\subsubsection{Centrality.}
The centrality of a node is defined as a number or rank corresponding to the node position within the network. 
By observing centrality, we can make considerations that allow us to better understand the process hidden in the ensemble method.
The notion of centrality encompasses a wide range of metrics. 
In this section, we explore those metrics that offer the most insightful information.

According to \citet{brandes_variants_2008}, we define the \emph{betweenness centrality} (BC) of a node as the fraction of all the shortest paths between every pair of nodes of the graph passing through the considered node. 
Let DPG = ($\mathcal{P}, E$) be the graph and $s, t, v\in\mathcal{P}$ three vertexes of DPG, we can denote with $\sigma(s,t)$ the number of shortest paths between $s$ and $t$ and with $\sigma(s,t | v)$ the number of shortest paths between $s$ and $t$ passing through $v$.
Then, the BC of the node $v\in\mathcal{P}$ is defined as:
\begin{equation*}
    BC(v) = \sum_{s,t\in\mathcal{P}}\frac{\sigma(s,t|v)}{\sigma(s,t)}.
\end{equation*}
All details and observations about BC can be read in \cite{brandes_variants_2008}.
BC serves as a relevant metric for gaining a deeper insight into the significance of decisions within the ensemble model. We can observe that a node with a higher BC value has a more significant influence on the flow of information within the graph; nodes with high BC can be considered potential bottleneck nodes because they play a crucial role in facilitating interactions between different parts of the DPG.
For this reason, we can assert that these nodes are meaningful to understanding the tree-based ensemble models: in all tree base learners, the decision contained in the node is essential to classify the elements of the dataset. We highlight that the significance extends beyond the characteristic itself; it encompasses the value associated with it.

According to \citet{mones_hierarchy_2012}, we define the \emph{local reaching centrality} (LRC) of a node $v$ of the DPG as the proportion of other nodes reachable from node $v$ via outgoing edges.
% Without going into too much detail, which can be read in \cite{mones_hierarchy_2012}, LRC can be generalised to weighted graphs by measuring the average weight of a given directed path starting from node $v$. 
LRC can be generalised to weighted graphs by measuring the average weight of a given directed path starting from node $v$ (more details are available on \cite{mones_hierarchy_2012}). 
The LRC serves as a metric for assessing the importance of DPG's nodes. 
It gauges the extent to which decisions contained in these nodes are employed by diverse tree base learners for classifying samples in the training set. 
This, in turn, reflects the importance of these decisions in the classification of new samples.
The LRC offers a comprehensive perspective on the concept of feature importance (FI) by extending its definition to encompass the values associated with features across various decisions. 
Additionally, the prominence of paths between highlighted predicates indicates their frequent utilisation, providing insights into how new samples can be classified with fewer decisions.

\subsubsection{Community.}
While there is no single definition of a community, we can observe structures similar to communities in the DPG. According to \citet{radicchi_defining_2004}, we define a \emph{community} as a subset of nodes of the DPG characterised by dense interconnections between its elements and sparse connections with the other nodes of the DPG that do not belong to the community.
Based on the properties of DPG, we employed \emph{asynchronous label propagation} algorithm, proposed by \citet{lpa_raghavan_near_2007}, to detect communities within the graph.
The core concept of the algorithm involves each graph node determining its community membership based on the majority of its neighbours.
Without delving into details, which can be appreciated in \cite{lpa_raghavan_near_2007}, the algorithm comprises a series of steps: each node initially possesses a unique label.
As these labels diffuse through the graph, closely connected groups of nodes converge on a common label. 
These consensus groups then expand outward until further expansion becomes impractical.
After this label propagation process, nodes sharing the same labels are identified as belonging to the same community. 
This process is iterated until each node in the network aligns its label with the community that includes the maximum number of its neighbouring nodes.
The algorithm is defined asynchronous, as each node receives updates without waiting for updates on the remaining nodes.
Identifying communities in the DPG provides insights into the ensemble model: visualising them allows us to discover groups of nodes that similarly contribute to the classification of samples.

By employing the asynchronous label propagation algorithm, we observe that each formed community is associated with a class.
Within these communities, the features utilised by the ensemble model to classify samples belonging to the community's class are emphasised. Once again, the association between features and values plays a key role, highlighting the specific decisions made by the learners.
To quote \citet{lpa_raghavan_near_2007}:
\begin{displayquote}
Communities in social networks can provide insights about common characteristics or beliefs among people that make them different from other communities.
\end{displayquote}
Similarly, we observe that communities within the DPG offer a valuable understanding of the characteristics for samples to be assigned to a particular community class. 
This intuition extends to identifying predominant features and those that play a marginal role in the classification process.
Moreover, it is noteworthy that communities also provide insights into the entire dataset and the complexity of the problem. 
A community comprising a substantial number of nodes, each associated with different predicates often involving distinct features, indicates that the model makes diverse decisions to assign samples to the community class. 
This implies that the model encounters challenges in classifying samples for this particular class, and data from different classes are not easily distinguishable within the dataset.
% We can note that, in a similar manner, the communities within DPG identify the precise characteristics that a sample must possess to be considered a candidate belonging to a specific class.
% This observation is particularly significant for gaining insights into the characteristics of the proposed dataset.
% Furthermore, as mentioned in the \emph{Cut Vertices} paragraph, frequently happens that classes are only a few nodes apart from each other. Communities provide an overview of what these nodes are, reducing the complexity of the problem.

Finally, communities, functioning as sub-graphs, can be used to visualise the decisions made in the ensemble model, enabling the identification of a specific class. This replaces the complex illustration of the DPG, especially when we are focused on visualising a single class and dealing with many tree base learners. We summarised the utility of discussed properties and metrics in \Cref{tab:resume}.
\begin{table}[h!]
\centering
\caption{Summary of Constraints, Betweenness Centrality (BC), Local Reaching Centrality (LRC), and Community, featuring provided definitions and their utility in offering insights into tree-based ensemble models.}
\label{tab:resume}
\begin{tabular}{p{1.7 cm}|p{4.7 cm}|p{5.5 cm}} %11.9
\toprule
\textbf{Property} & \textbf{Definition} & \textbf{Utility} \\
\midrule
Constraints &
The intervals of values for each feature obtained from all predicates connected by a path that culminates in a given class.
&
Calculate the classification boundary values of each feature associated with each class.
\\
\midrule
BC &
%Measure that quantifies the fraction of all the shortest paths between every pair of nodes of the graph passing through the considered node 
Quantifies the fraction of all the shortest paths between every pair of nodes of the graph passing through the considered node.
&
Identify potential bottleneck nodes that correspond to crucial decisions.
%made by many tree base learners in the classification process
\\
\midrule
LRC &
%Measure that quantifies the proportion of other nodes reachable from the focal node through its outgoing edges 
Quantifies the proportion of other nodes reachable from the local node through its outgoing edges.
&
%Assess the importance of nodes, which reflects the significance of the decisions in the classification process. Extend the definition of feature importance encompassing the values associated with features across the decisions
Assess the importance of nodes similarly to feature importance, but enrich the information by encompassing the values associated with features across all decisions.
\\
\midrule
Community &
A subset of nodes of the DPG which is characterised by dense interconnections between its elements and sparse connections with the other nodes of the DPG that do not belong to the community.
&
Understanding the characteristics of nodes to be assigned to a particular community class, identifying predominant predicates, and those that play a marginal role in the classification process.
%A high number of nodes involved in the same community implies that the model encounters challenges in classifying samples for that particular class, and data from different classes are not easily distinguishable within the dataset
\\
\bottomrule
\end{tabular}
\end{table}

\section{Empirical Results and Discussion}
%Specify the datasets used for evaluation.
%Describe the tree-based ensemble models employed in the experiments.
%Outline the metrics used to assess the effectiveness of the mapping approach.

%Present the results of the experiments, including the mapped DPGs and their interpretations.
%Discuss the insights gained from the mapped DPGs and their relevance to model understanding and explanation in XAI.
%Compare the performance of the mapped DPGs with other interpretation methods, if applicable.
%Intro sezione

In this section, we demonstrate the effectiveness of DPG to the well-known Iris dataset \cite{fisher1936use} and a synthetic multiclass dataset\footnote{\url{https://github.com/sbarbonjr/fhg/tree/main/datasets}}. Each experiment in this section was conducted using the RF classifier, with variations limited to the number of tree base learners. The implementation is available here\footnote{\url{https://github.com/sbarbonjr/fhg}}. %, and a random seed of $42$ was utilised.
Finally, we discuss potential enhancements to DPG and explore further development opportunities.

\subsection{DPG: Iris insights}\label{emp_overview}
The first case study concerns the classification of the Iris dataset \cite{fisher1936use}.
The simplicity, manageability, versatility, and relevance of this dataset make it an interesting and relevant resource for discussions and demonstrations of interpretability in ML. 
The dataset comprises measurements of sepals and petals for iris flowers encompassing three distinct species with a total of four features and three classes.
The RF was selected as the tree-based ensemble model due to its well-established reputation and high-performance capabilities.
% The dataset was split into a \num{70}\% train set and a \num{30}\% test set, employing stratification based on the target variable.
To conduct the classification, we partitioned the dataset into training and test sets, following a \num{80}-\num{20}\% proportion, respectively. A seed value of \num{42} was established for randomness control, and the number of base tree learners was set to \num{5}.
The RF performances, evaluated on the test set, are summarised in the confusion matrix in \Cref{tab:cm_5_bl}. The model demonstrates \num{100}\% accuracy.
\begin{table}[htbp!]
\centering
\caption{Confusion matrix depicting the performance evaluation of the RF model with \num{5} base tree learners on the test set.}
\label{tab:cm_5_bl}
\begin{tabular}{c @{\hskip 0.15cm}| c c c}
            \toprule
             & \multicolumn{3}{c}{\textbf{Prediction}} \\
            \textbf{Ground truth} & Class 0 & Class 1 & Class 2 \\
            \midrule
            Class 0 & 19 & 0 & 0 \\
            Class 1 & 0 & 13 & 0 \\
            Class 2 & 0 & 0 & 13 \\
            \bottomrule
        \end{tabular}
\end{table}

\begin{figure}[t]
\centering
\includegraphics[width=\columnwidth]{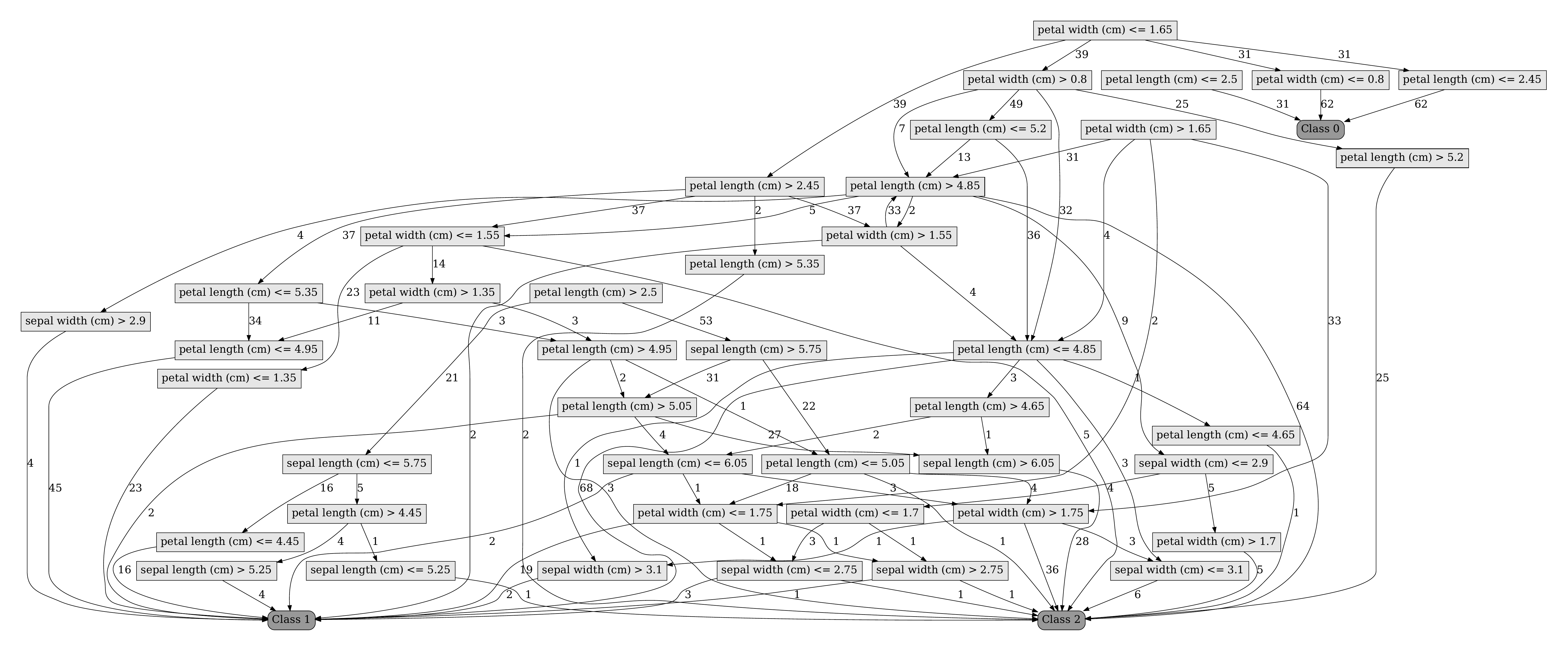}
\caption{DPG of the RF composed of \num{5} tree base learners trained on Iris dataset.}\label{fig:iris_bl2_DPG}
\end{figure}

After training the model, we applied the algorithm outlined in \Cref{sec:pseudo_code} to obtain the DPG, which can be visualised in \Cref{fig:iris_bl2_DPG}.
Then, we can analyse the obtained graph using the metrics and algorithms proposed in \Cref{DPG_inter}, verifying their utility and effectiveness in gaining insights into the model.
It is important to note that the DPG leads to the calculation of both global metrics, referring to the overall graph, and metrics at the level of individual nodes.
% By interpreting both types of metrics, we can gain insights into the most relevant predicates or features, the presence or absence of communities, and the paths most frequently utilised by the samples.

To illustrate the effectiveness and one of the advantages of employing DPG, we highlight the constraints for the different classes in the \Cref{tab:constraints}. 
The class-specific constraints delineate the necessary characteristics a sample must exhibit to be assigned to that particular class by the tree-based ensemble model. 
This insight contributes to a better understanding of how the model utilises features for effective classification.

\begin{table}[htbp!]  
\centering
\caption{Constraints for each class based on the DPG for an RF model within \num{5} tree base learners.}
\label{tab:constraints}
\begin{tabular}{c|p{5.cm}}
\toprule
\textbf{Class} & \textbf{Constraints} \\
\midrule
\multirow{2}*{0} & \texttt{petal width (cm)} $\leq$ 1.65 \\ 
                 & \texttt{petal length (cm)} $\leq$ 2.50 \\
\midrule
\multirow{4}*{1} & 5.25 < \texttt{sepal length (cm)} $\leq$ 6.05 \\ 
                 & 0.80 < \texttt{petal width (cm)} $\leq$ 1.75 \\ 
                 & 2.45 < \texttt{petal length (cm)} $\leq$ 5.35 \\ 
                 & 2.75 < \texttt{sepal width (cm)} $\leq$ 2.90 \\
\midrule
\multirow{4}*{2} & 5.75 < \texttt{sepal length (cm)} $\leq$ 6.05 \\
                 & 0.80 < \texttt{petal width (cm)} $\leq$ 1.75 \\ 
                 & 2.45 < \texttt{petal length (cm)} $\leq$ 5.35 \\ 
                 & 2.75 < \texttt{sepal width (cm)} $\leq$ 3.10 \\
\bottomrule
\end{tabular}
\end{table}

%The first metric under discussion is the presence of cut-vertices in DPG. 
%We promptly observe that \texttt{petal width (cm) <= 1.65} is a cut-vertex. 
%Upon its removal, two connected components emerge: one exclusively containing \emph{Class 0} and the other encompassing the remaining two classes.

The first metric under discussion is the BC of the nodes, as depicted in \Cref{tab:bc}, where we identify potential bottleneck nodes. 
These nodes encapsulate significant information, particularly representing decisions made by numerous tree base learners. 
We quickly discern that the decision associated with \texttt{petal length (cm)} and the value \num{4.85} is pivotal, as it is frequently relied upon by multiple basic learners and is essential for successful classification.
\begin{table}[t!]
\centering
\caption{Top eight predicates by evaluating their BC obtained from the DPG based on an RF model consisting of \num{5} tree base learners.}
    \label{tab:bc}
    \begin{tabular}{l | c}
    \toprule
    \textbf{Predicate} & \textbf{BC} \\
    \midrule
        \texttt{petal length (cm) > 4.85} & 0.053 \\
	\texttt{petal length (cm) <= 4.85} & 0.036 \\
	\texttt{petal width (cm) > 1.55} & 0.034 \\
	\texttt{sepal length (cm) <= 6.05} & 0.032 \\
	\texttt{petal length (cm) > 4.95} & 0.028 \\
	\texttt{petal length (cm) > 4.65} & 0.022 \\
	\texttt{petal width (cm) <= 1.75} & 0.022 \\
	\texttt{petal width (cm) <= 1.55} & 0.021 \\
    \bottomrule
    \end{tabular}
\end{table}

Furthermore, the LRC metric provides additional information. 
Examining \Cref{subtab:metrics}, we can emphasise the most crucial predicates influencing the decision-making process of the tree-based ensemble model. 
This includes not only identifying the most frequently used features but also recognising the associated values that lead to significant and divisive splits in the dataset across various basic learners. 
As observed in \Cref{tab:lrc}, a comparison between the LRC of the nodes and the FI, calculated on the same model on which DPG is based, suggests that the metric may rank the predicates similarly. 
FI is calculated using the \emph{Mean Decrease Impurity} (MDI) algorithm introduced by \citet{scornet_trees_2021}. 
This comparison also provides additional information about the values used in the decisions and the frequency of paths extending the concept of FI.

\begin{table}[htbp!]
    \caption{Comparison of the top eight predicates by evaluating their LRC obtained from the DPG based on an RF model consisting of \num{5} tree base learners (\Cref{subtab:metrics}), alongside the FI of the same model (\Cref{subtab:fi}) calculated exploiting MDI algorithm.}
    \label{tab:lrc}
    \begin{subtable}{.5\linewidth}
      \centering
        \caption{LRC evaluation}
            \begin{tabular}{l | c}
            \toprule
            \textbf{Predicate} & \textbf{LRC} \\
            \midrule
            \texttt{petal width (cm) <= 1.65} & 1.531 \\
            \texttt{petal length (cm) > 2.45} & 0.919 \\
            \texttt{petal width (cm) > 0.80} & 0.874 \\
            \texttt{petal length (cm) > 2.50} & 0.699 \\
            \texttt{petal width (cm) > 1.65} & 0.618 \\
            \texttt{petal length (cm) <= 5.20} & 0.565 \\
            \texttt{petal width (cm) > 1.55} & 0.540 \\
            \texttt{sepal length (cm) > 5.75} & 0.332 \\
            \bottomrule
            \end{tabular}
        \label{subtab:metrics}
    \end{subtable}%
    \begin{subtable}{.5\linewidth}
      \centering
        \caption{FI evaluation}
        \begin{tabular}{l | c}
            \toprule
            \textbf{Feature} & \textbf{FI} \\
            \midrule
            \texttt{petal length (cm)} & 0.550 \\
            \texttt{petal width (cm)} & 0.373 \\
            \texttt{sepal length (cm)} & 0.054 \\
            \texttt{sepal width (cm)} & 0.023 \\
            \bottomrule
            \end{tabular}
        \label{subtab:fi}
    \end{subtable} 
\end{table}

By employing the global metric community, we identified the presence of three distinct communities. \Cref{tab:communities} illustrates the association between each community and a distinct class obtained by applying the asynchronous label propagation algorithm to the DPG. We can affirm that each node within a community contains decisions that significantly contribute to the accurate classification of samples belonging to a specific class. For instance, when applying the predicates of Community 3 (comprising two features and two predicates) to the test set and traversing from the root node, it achieved \num{100}\% accuracy for Class 0, the class delineated in the mentioned community.
\begin{table}[h!]
\centering
\caption{Communities obtained from the DPG based on an RF model composed of \num{5} tree base learners. The table shows the number of predicates belonging to each community, the number of features in the community nodes, and the class involved in each community.}
\label{tab:communities}
\begin{tabular}{r|c|c|c}
\toprule
\textbf{Community} & \textbf{\# Predicates} & \textbf{\# Features} & \textbf{Class} \\
\midrule
Community 1 & 23 & 4 & 1 \\
\midrule
Community 2 & 18 & 4 & 2 \\
\midrule
Community 3 & 3 & 2 & 0 \\
\bottomrule
\end{tabular}
\end{table}

\begin{figure}[htbp!]
  \centering
  \includegraphics[width=\textwidth]{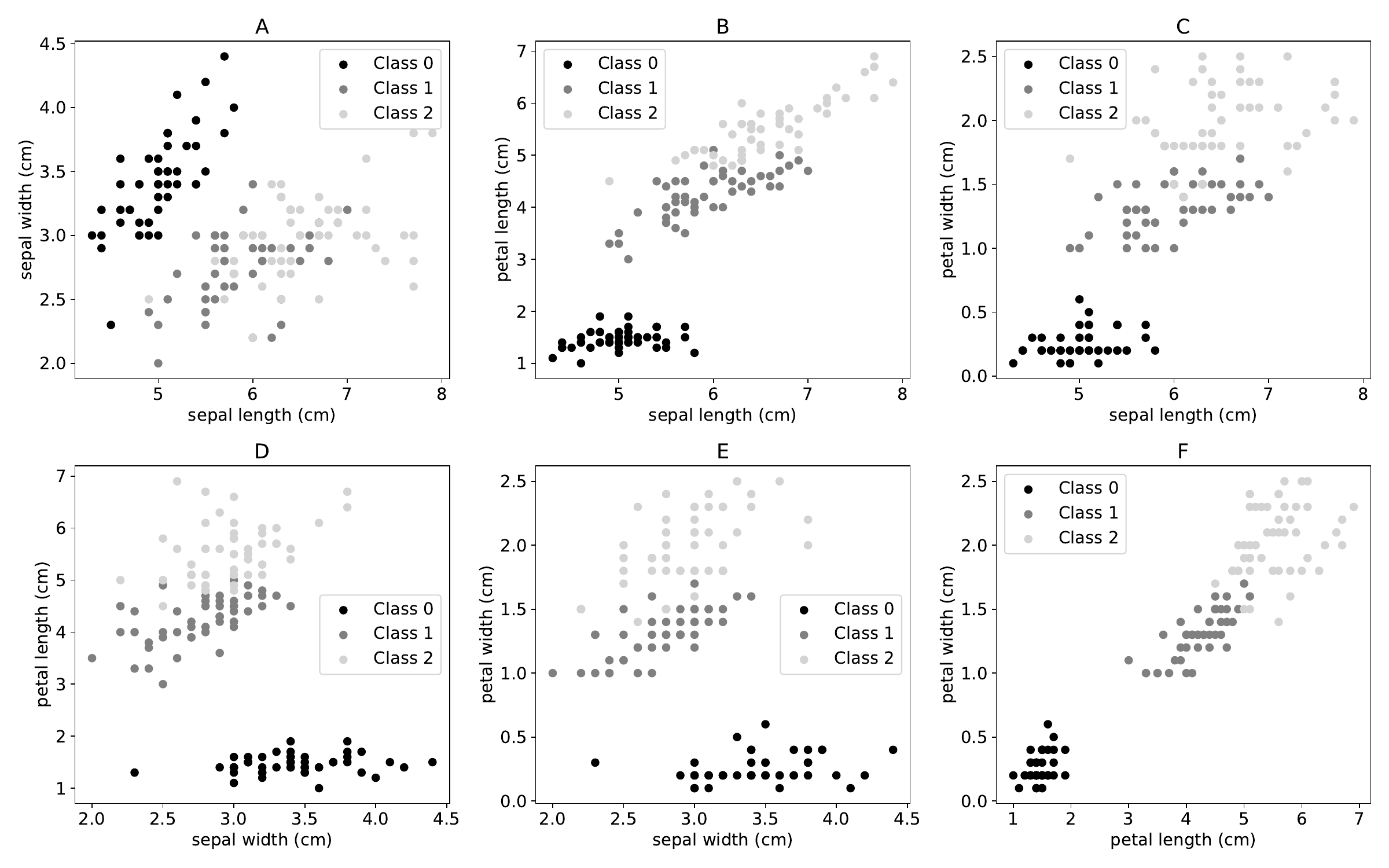}
  \caption{Two-dimensional depiction of the Iris dataset, employing feature pairs in each graph for visual representation.}
  \label{fig:iris_plots}
\end{figure}

Finally, upon comparing the \Cref{tab:communities} and the \Cref{fig:iris_plots}, we can state that the communities facilitate the comprehension of how the model addresses the classification problem.
Examining the number of decisions and features utilised in each community reveals that differentiating between Class 1 (Community 1) and Class 2 (Community 2) poses a greater challenge for the model. 
This indicates the difficulty in effectively separating samples within the dataset into their respective classes.
This difficulty becomes apparent when visualising the dataset across the features, as in \Cref{fig:iris_plots}.
Conversely, the community encompassing Class 0 (Community 3) consists of fewer predicates, signifying that it is more distinguishable from other classes, as confirmed in the \Cref{fig:iris_plots}.

\subsection{Comparing to the Graph-based Solutions}

\begin{figure}[t!]
\centering
\includegraphics[width=\columnwidth]{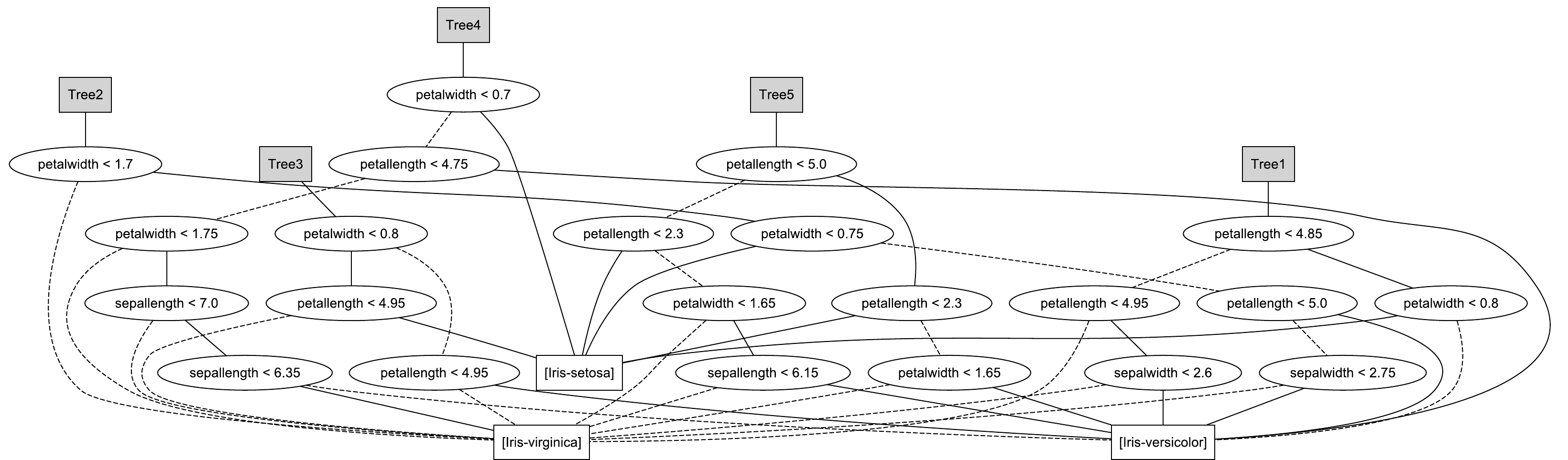}
\caption{ADD of an RF model with \num{5} tree base learners induced for Iris dataset.}\label{fig:iris_bl5_ADD}
\end{figure}

As outlined in the \Cref{LR}, \citet{Gossen2023267,murtovi_forest_2023} conducted studies on the interpretability of tree-based ensemble models exploiting graph structures.
We compared DPG with their proposed method by examining their outcomes and potential insights.
Using the same tree-based ensemble model we studied in \Cref{emp_overview} as input, we generated the ADD displayed in \Cref{fig:iris_bl5_ADD}. 
The first noticeable distinction from DPG lies in the ADD structure, as it forms a bipartite graph.
We can observe that the ADD is generated from the trained model, albeit without fully leveraging the training dataset.
Consequently, the evaluation of connections between nodes and the assessment of the significance of decisions made by different tree base learners are not fully exploited.
This implies that each branch carries equal weight and impact in the diagram, and a classification error significantly influences the structure by introducing an incorrect path.
Another crucial difference is that ADD does not provide graph metrics, leaving the user the interpretation of the diagram and potentially missing out on relevant information.
Moreover, an additional limitation, as indicated in the studies by \cite{Gossen2023267, murtovi_forest_2023}, involves the challenge that emerges when generating ADDs and dealing with large ensembles. 
Visualisation becomes intricate, even with a modest count of \num{20} tree base learners.
In contrast, DPG allows the computation of both global and local metrics, even with a higher number of tree base learners.

To further examine these aspects, we use two RF models, one with \num{20} tree learners and the other with \num{100}, to analyse a complex multiclass problem with a dataset comprising \num{4} classes, \num{1000} samples, and \num{16} features.
% , of which \num{7} are deemed significant.
This introduces a four-class problem that was randomly generated.
The dataset was created using the \texttt{make\_classification}\footnote{\url{https://scikit-learn.org/stable/modules/generated/sklearn.datasets.make\_classification.html}} function from \texttt{scikit-learn}.
The following setup has been maintained for training both models.
We divided the dataset into training and test sets, with an \num{80}-\num{20}\% ratio. 
We fixed a seed value of \num{42} for randomness control to ensure reproducibility. 
The RF performances, assessed on the test set, are summarised in the confusion matrices presented in \Cref{tab:cm_bls}. 
The model with \num{100} tree base learners shows better performance in every parameter, reaching an overall accuracy of \num{66}\%, outperforming the model with \num{20} tree base learners, which barely reaches \num{58}\%.
\begin{table}[h!]
\centering
\caption{Confusion matrices of the RF models with \num{20} tree base learners (RF 20) and with \num{100} tree base learners (RF 100) tested on the synthetic dataset.}
\label{tab:cm_bls}
\begin{tabular}{c|c c c c|c c c c}
\toprule
& \multicolumn{4}{c}{\textbf{Prediction (RF 20)}} & \multicolumn{4}{c}{\textbf{Prediction (RF 100)}} \\
\textbf{Ground truth} & Class 0 & Class 1 & Class 2 & Class 3 & Class 0 & Class 1 & Class 2 & Class 3\\
\midrule
Class 0 & 38 & 4 & 12 & 6 & 38 & 4 & 14 & 4 \\
Class 1 & 5 & 31 & 3 & 5 & 4 & 32 & 2 & 6 \\
Class 2 & 11 & 2 & 29 & 2 & 5 & 2 & 33 & 4 \\
Class 3 & 10 & 13 & 10 & 19 & 9 & 9 & 5 & 29  \\
\bottomrule
\end{tabular}
\end{table}

We emphasise that even in this context, DPG is a useful tool. 
Both DPG and ADD present intricate visualisations with \num{20} tree base learners. 
However, DPG overcomes this obstacle by providing metrics that can still offer valid insights into the model.
The first insight is displayed in \Cref{tab:constraints_bl100}, where we provide constraints for the four classes of the dataset.
Constraints, even in this complex scenario, allow the visualisation of intervals where sample features should be situated for precise classification into their respective classes.
% The comprehensive table containing all constraints and their corresponding classes is available in the repository.

\begin{table}[htbp!]
\centering
 \caption{Constraints for each class based on the DPG for an RF model within \num{100} tree base learners.}
\label{tab:constraints_bl100}
\begin{tabular}{c|c|c|c}
 \toprule
    Class 0 & Class 1 & Class 2 & Class 3 \\
    \midrule
    $-5.87 < \texttt{F1} \leq 5.74$ & $-5.79 < \texttt{F1} \leq 5.72$ & $-5.76 < \texttt{F1} \leq 5.72$ & $-5.76 < \texttt{F1} \leq 5.72$ \\
    $-2.64 < \texttt{F2} \leq 2.63$ & $-2.61 < \texttt{F2} \leq 2.61$ & $-2.61 < \texttt{F2} \leq 2.61$ & $-2.61 < \texttt{F2} \leq 2.63$ \\
    $-5.24 < \texttt{F3} \leq 3.75$ & $-5.38 < \texttt{F3} \leq 3.75$ & $-5.24 < \texttt{F3} \leq 3.75$ & $-5.24 < \texttt{F3} \leq 3.75$ \\
    $-4.80 < \texttt{F4} \leq 4.37$ & $-5.15 < \texttt{F4} \leq 4.37$ & $-4.80 < \texttt{F4} \leq 4.37$ & $-4.80 < \texttt{F4} \leq 4.37$ \\
    $-2.61 < \texttt{F5} \leq 2.44$ & $-2.61 < \texttt{F5} \leq 2.71$ & $-2.61 < \texttt{F5} \leq 2.44$ & $-2.61 < \texttt{F5} \leq 2.44$ \\
    $-2.29 < \texttt{F6} \leq 2.58$ & $-2.29 < \texttt{F6} \leq 2.58$ & $-2.29 < \texttt{F6} \leq 2.58$ & $-2.29 < \texttt{F6} \leq 2.58$ \\
    $-2.82 < \texttt{F7} \leq 2.47$ & $-2.82 < \texttt{F7} \leq 2.47$ & $-2.82 < \texttt{F7} \leq 2.47$ & $-2.82 < \texttt{F7} \leq 2.86$ \\
    $-4.62 < \texttt{F8} \leq 4.74$ & $-4.62 < \texttt{F8} \leq 4.74$ & $-4.62 < \texttt{F8} \leq 4.74$ & $-4.62 < \texttt{F8} \leq 4.74$ \\
    $-2.40 < \texttt{F9} \leq 2.59$ & $-2.40 < \texttt{F9} \leq 2.59$ & $-2.40 < \texttt{F9} \leq 2.71$ & $-2.40 < \texttt{F9} \leq 2.59$ \\
    $-4.71 < \texttt{F10} \leq 4.32$ & $-4.71 < \texttt{F10} \leq 4.32$ & $-4.71 < \texttt{F10} \leq 5.43$ & $-4.71 < \texttt{F10} \leq 4.32$ \\
    $-2.87 < \texttt{F11} \leq 2.77$ & $-2.87 < \texttt{F11} \leq 2.77$ & $-2.87 < \texttt{F11} \leq 2.86$ & $-2.87 < \texttt{F11} \leq 2.77$ \\
    $-2.42 < \texttt{F12} \leq 2.37$ & $-2.42 < \texttt{F12} \leq 2.37$ & $-2.42 < \texttt{F12} \leq 2.37$ & $-2.42 < \texttt{F12} \leq 2.37$ \\
    $-4.28 < \texttt{F13} \leq 5.01$ & $-4.28 < \texttt{F13} \leq 5.01$ & $-4.28 < \texttt{F13} \leq 5.01$ & $-4.28 < \texttt{F13} \leq 5.01$ \\
    $-7.31 < \texttt{F14} \leq 8.33$ & $-7.31 < \texttt{F14} \leq 8.33$ & $-7.31 < \texttt{F14} \leq 8.33$ & $-7.31 < \texttt{F14} \leq 8.33$ \\
    $-4.46 < \texttt{F15} \leq 4.19$ & $-4.46 < \texttt{F15} \leq 4.19$ & $-4.46 < \texttt{F15} \leq 4.19$ & $-4.97 < \texttt{F15} \leq 4.5$ \\
    $-4.70 < \texttt{F16} \leq 4.21$ & $-4.70 < \texttt{F16} \leq 4.21$ & $-4.70 < \texttt{F16} \leq 4.52$ & $-4.70 < \texttt{F16} \leq 4.21$ \\
    \bottomrule
\end{tabular}
\end{table}

% \begin{figure}[t!] 
% \caption{Confusion matrices of the RF models with \num{20} tree base learners (\Cref{tab:cm_20_bl}) and with \num{100} tree base learners (\Cref{tab:cm_100_bl}) tested on the synthetic dataset.}
%   \begin{subtable}{0.5\textwidth}
%     \centering
%     \caption{\num{20} tree base learners}
%     \label{tab:cm_20_bl}
%     \begin{tabular}{r @{\hskip 0.15cm} c c c c}
%       \cmidrule{2-5}
%       & \multicolumn{4}{c}{\textbf{Prediction}} \\
%       \textbf{Ground truth} & Class 0 & Class 1 & Class 2 & Class 3 \\
%       \cmidrule{2-5}
%       Class 0 & 38 & 4 & 12 & 6\\
%       Class 1 & 5 & 31 & 3 & 5 \\
%       Class 2 & 11 & 2 & 29 & 2 \\
%       Class 3 & 10 & 13 & 10 & 19 \\
%       \cmidrule{2-5}
%     \end{tabular}
%   \end{subtable}%
%   \begin{subtable}{0.5\textwidth}
%     \centering
%     \caption{\num{100} tree base learners}
%     \label{tab:cm_100_bl}
%     \begin{tabular}{r @{\hskip 0.15cm} c c c c}
%       \cmidrule{2-5}
%       & \multicolumn{4}{c}{\textbf{Prediction}} \\
%         \textbf{Ground truth} & Class 0 & Class 1 & Class 2 & Class 3 \\
%       \cmidrule{2-5}
%        Class 0 & 38 & 4 & 14 & 4\\
%        Class 1 & 4 & 32 & 2 & 6 \\
%        Class 2 & 5 & 2 & 33 & 4 \\
%        Class 3 & 9 & 9 & 5 & 29 \\
%       \cmidrule{2-5}
%     \end{tabular}
%   \end{subtable}
%   \label{fig:cm_bls}
% \end{figure}

The BC metric helps identifying potential bottleneck nodes.
Upon observing \Cref{tab:bc_100BL}, we can see that there is not a large difference between the BC values associated with the predicates.
Therefore, we can assume that there are no bottleneck nodes.

Examining the information provided in \Cref{tab:lrc_100BL}, the LRC underscores which predicates significantly impact the decision-making process of the ensemble model. 
The predicates \texttt{F7 <= 1.62} and \texttt{F1 <= \num{3.1}} are identified as particularly crucial.

\begin{table}[ht!]
\centering
\caption{Top eight predicates by evaluating their BC (\Cref{tab:bc_100BL}), and top eight predicates by evaluating their LRC (\Cref{tab:lrc_100BL}), both obtained from the DPG based on an RF model consisting of \num{100} tree base learners.}
    \label{tab:lrcbc}
    \begin{subtable}{.5\linewidth}
      \centering
        \caption{BC evaluation}
            \begin{tabular}{l | c}
            \toprule
            \textbf{Predicate} & \textbf{BC} \\
            \midrule
            \texttt{F15 > 1.17} & 0.018 \\
            \texttt{F15 <= 1.61} & 0.015 \\
            \texttt{F12 > 0.20}	& 0.015 \\
            \texttt{F12 > 0.41}	& 0.014 \\
            \texttt{F4 <= 0.33}	& 0.014 \\
            \texttt{F8 > 0.36} & 0.014 \\
            \texttt{F1 <= -1.10} & 0.013 \\
            \texttt{F11 <= 1.16} & 0.013 \\
            \bottomrule
            \end{tabular}
        \label{tab:bc_100BL}
    \end{subtable}%
    \begin{subtable}{.5\linewidth}
      \centering
        \caption{LRC evaluation}
        \begin{tabular}{l | c}
            \toprule
            \textbf{Predicate} & \textbf{LRC} \\
            \midrule
            \texttt{F7 <= 1.62} & 15.812 \\
            \texttt{F1 <= 3.10}	& 14.475 \\
            \texttt{F14 > -1.78} & 13.313 \\
            \texttt{F4 > -2.97} & 13.158 \\
            \texttt{F5 > -1.92} & 13.065 \\
            \texttt{F4 > -1.36} & 12.989 \\
            \texttt{F1 <= 2.49} & 12.986 \\
            \texttt{F13 > 1.98} & 12.920 \\
            \bottomrule
            \end{tabular}
        \label{tab:lrc_100BL}
    \end{subtable} 
\end{table}

% \begin{figure}[t!]
%   \centering
%   \caption{Comparison of the features importance.}
  
%   \begin{subfigure}{0.48\textwidth}
%     \includegraphics[width=\linewidth]{Pictures/FI_LR.png} 
%     %\caption{Confusion matrix for 20 tree base learners}
%     \label{fig:FI_LR}
%   \end{subfigure}\hfill
%   \begin{subfigure}{0.48\textwidth}
%     \includegraphics[width=\linewidth]{Pictures/FI_MDI.png} 
%     %\caption{Confusion matrix for 100 tree base learners}
%     \label{fig:FI_MDI}
%   \end{subfigure}
  
%   \label{fig:cm_fi}
% \end{figure}

Another insight can be obtained by employing the global metric community.
In this scenario, we identified the presence of four distinct communities, displayed in \Cref{tab:communities_bl100}.
We note that each community contains a distinct class.
Furthermore, upon observing the table, we can conclude that each community exhibits a high number of involved features and predicates, confirming the complexity of the classification problem.

\begin{table}[h!]
\centering
\caption{Communities obtained from an RF model composed of \num{100} tree base learners. The table shows the number of predicates belonging to each community, the number of features in the community nodes, and the class involved in each community.}
\label{tab:communities_bl100}
\begin{tabular}{r|c|c|c}
\toprule
\textbf{Community} & \textbf{\# Predicates} & \textbf{\# Features} & \textbf{Class} \\
\midrule
Community 1 & 7767 & 16 & 2 \\
\midrule
Community 2 & 2149 & 16 & 0 \\
\midrule
Community 3 & 2351 & 16 & 3 \\
\midrule
Community 3 & 2100 & 16 & 1 \\
\bottomrule
\end{tabular}
\end{table}

%The obtained results align with the details provided in the \ref{DPG_inter} section and are extensively documented within the paper's GitHub repository.

% \begin{figure}[b]
%   \centering
%   \includegraphics[width=.2\textwidth]{Pictures/new_bl5_perc0.0_dec2_temp.pdf}
%   \caption{visualisation of NEW dataset XXXX}
%   \label{fig:iris_plots}
% \end{figure}

% \begin{figure}[htbp!]
%   \centering
%   \includegraphics[width=\textwidth]{Pictures/new_bl20_perc0.png}
%   \caption{visualisation of NEW dataset XXXX, 20 bl}
%   \label{fig:iris_plots}
% \end{figure}

% \begin{figure}[htbp!]
%   \centering
%   \includegraphics[width=\textwidth]{Pictures/pca_visualisation.pdf}
%   \caption{PCA visualisations of the new Dataset XXX}
%   \label{fig:iris_plots}
% \end{figure}

\subsection{Potential Improvements}
Several avenues await exploration in the future.
The primary aim is to reduce the computational cost of DPG, as many real-world problems involve large datasets that do not scale well with the current implementation of DPG.
% Given that DPG scales quadratically, it is imperative to identify methods for optimising its runtime.
Expanding the application scope of DPG is another key goal, including its utility in explaining models relevant to regression-type problems.
Given DPG's applicability to any model and dataset, we aim to introduce new tests and use cases to delve deeper into the method. 
This includes proposing applications to novel datasets and exploring their compatibility with other tree-based ensemble models.
Furthermore, while this paper introduces certain metrics and algorithms derived from graph theory, the field offers extensive possibilities for future exploration. In the future, we plan to introduce new tools associated with DPG to enhance the interpretation of tree-based ensemble models.

\section{Conclusion}
In this paper, we introduced Decision Predicate Graphs (DPG) as a novel model-agnostic tool for tree-based ensemble interpretability. 
%DPG is directly applied to the trained model, ensuring the maintenance of its performance.
DPG is obtained from a trained model and data, ensuring the maintenance of its performance. The concept behind DPG is to convert an opaque-box tree-based ensemble model into an enriched graph. DPG enables qualitative measures and the identification of its predicates, and facilitates comparisons between features and their associated values, offering insights into the entire model. In particular, we introduced Betweenness Centrality, Local Reaching Centrality, Community and Constraints as useful metrics and properties towards improving and extending the XAI interpretability approaches. While DPG is still considered an evolving work, its potential is substantial, given the robust underlying theory and the versatility of the tool. Furthermore, the effervescent research on graphs, knowledge graphs, and complex networks might strengthen the possibilities grounded in DPG. As the next step of our current research, we expect to apply DPG to provide local interpretability and also contribute to improving the interpretation of regression tasks.

%Moreover, DPG highlights connections between tree base learners, providing understandings into which features are most frequently used and how they contribute to classifying samples.

%We presented the formalisation of DPG and provided the algorithm, accompanied by pseudo-code, to help comprehension and reproducibility.
%Drawing on graph theory and the properties of DPG, we proposed valid metrics to gain insights into ensemble models. 
%In particular, we introduced centrality and community concepts.
% We conduct two case studies using random forests, varying the number of basic learners, on both the Iris dataset and a synthetic dataset.

%We conducted studies aimed to validate the effectiveness of DPG in different contexts of difficulty demonstrating the utility of the method and its effectiveness in different scenarios.

% However, there are still notable limitations, such as high computational complexity and challenges in visualising large models with a significant number of learners.
% Despite these limitations, 
%DPG proves to be a valuable tool for comprehending the intricate decisions made by tree-based ensemble models.

\clearpage

\bibliography{bibliography.bib}

\begin{thebibliography}{44}
\providecommand{\natexlab}[1]{#1}
\providecommand{\url}[1]{\texttt{#1}}
\expandafter\ifx\csname urlstyle\endcsname\relax
  \providecommand{\doi}[1]{doi: #1}\else
  \providecommand{\doi}{doi: \begingroup \urlstyle{rm}\Url}\fi

\bibitem[Malekloo et~al.()Malekloo, Ozer, {AlHamaydeh}, and Girolami]{malekloo_machine_2022}
Arman Malekloo, Ekin Ozer, Mohammad {AlHamaydeh}, and Mark Girolami.
\newblock Machine learning and structural health monitoring overview with emerging technology and high-dimensional data source highlights.
\newblock 21\penalty0 (4):\penalty0 1906--1955.
\newblock ISSN 1475-9217.
\newblock \doi{10.1177/14759217211036880}.
\newblock Publisher: {SAGE} Publications.

\bibitem[Mienye and Sun()]{mienye_survey_2022}
Ibomoiye~Domor Mienye and Yanxia Sun.
\newblock A survey of ensemble learning: Concepts, algorithms, applications, and prospects.
\newblock 10:\penalty0 99129--99149.
\newblock ISSN 2169-3536.
\newblock \doi{10.1109/ACCESS.2022.3207287}.
\newblock Conference Name: {IEEE} Access.

\bibitem[Hastie et~al.()Hastie, Tibshirani, and Friedman]{hastie_additive_2009}
Trevor Hastie, Robert Tibshirani, and Jerome Friedman.
\newblock Additive models, trees, and related methods.
\newblock In Trevor Hastie, Robert Tibshirani, and Jerome Friedman, editors, \emph{The Elements of Statistical Learning: Data Mining, Inference, and Prediction}, Springer Series in Statistics, pages 295--336. Springer.

\bibitem[Chimatapu et~al.(2018)Chimatapu, Hagras, Starkey, and Owusu]{chimatapu2018explainable}
Ravikiran Chimatapu, Hani Hagras, Andrew Starkey, and Gilbert Owusu.
\newblock Explainable ai and fuzzy logic systems.
\newblock In \emph{Theory and Practice of Natural Computing: 7th International Conference, TPNC 2018, Dublin, Ireland, December 12--14, 2018, Proceedings 7}, pages 3--20. Springer, 2018.

\bibitem[Ho()]{ho_random_1995}
Tin~Kam Ho.
\newblock Random decision forests.
\newblock In \emph{Proceedings of 3rd International Conference on Document Analysis and Recognition}, volume~1, pages 278--282 vol.1.
\newblock \doi{10.1109/ICDAR.1995.598994}.

\bibitem[Breiman()]{breiman_random_2001}
Leo Breiman.
\newblock Random forests.
\newblock 45\penalty0 (1):\penalty0 5--32.
\newblock ISSN 1573-0565.
\newblock \doi{10.1023/A:1010933404324}.

\bibitem[Gossen and Steffen()]{Gossen2023267}
Frederik Gossen and Bernhard Steffen.
\newblock Algebraic aggregation of random forests: towards explainability and rapid evaluation.
\newblock 25\penalty0 (3):\penalty0 267--285.
\newblock ISSN 1433-2787.
\newblock \doi{10.1007/s10009-021-00635-x}.

\bibitem[Fisher(1936)]{fisher1936use}
Ronald~A Fisher.
\newblock The use of multiple measurements in taxonomic problems.
\newblock \emph{Annals of eugenics}, 7\penalty0 (2):\penalty0 179--188, 1936.

\bibitem[Guidotti et~al.(2018)Guidotti, Monreale, Ruggieri, Turini, Giannotti, and Pedreschi]{guidotti2018survey}
Riccardo Guidotti, Anna Monreale, Salvatore Ruggieri, Franco Turini, Fosca Giannotti, and Dino Pedreschi.
\newblock A survey of methods for explaining black box models.
\newblock \emph{ACM Comput. Surv.}, 51\penalty0 (5), aug 2018.
\newblock ISSN 0360-0300.
\newblock \doi{10.1145/3236009}.

\bibitem[Hanif et~al.()Hanif, Zhang, and Wood]{hanif2021survey}
Ambreen Hanif, Xuyun Zhang, and Steven Wood.
\newblock A survey on explainable artificial intelligence techniques and challenges.
\newblock In \emph{2021 {IEEE} 25th International Enterprise Distributed Object Computing Workshop ({EDOCW})}, pages 81--89.
\newblock \doi{10.1109/EDOCW52865.2021.00036}.
\newblock {ISSN}: 2325-6605.

\bibitem[Adadi and Berrada()]{adadi2018peeking}
Amina Adadi and Mohammed Berrada.
\newblock Peeking inside the black-box: A survey on explainable artificial intelligence ({XAI}).
\newblock 6:\penalty0 52138--52160.
\newblock ISSN 2169-3536.
\newblock \doi{10.1109/ACCESS.2018.2870052}.
\newblock Conference Name: {IEEE} Access.

\bibitem[Dwivedi et~al.()Dwivedi, Dave, Naik, Singhal, Omer, Patel, Qian, Wen, Shah, Morgan, and Ranjan]{Dwivedi}
Rudresh Dwivedi, Devam Dave, Het Naik, Smiti Singhal, Rana Omer, Pankesh Patel, Bin Qian, Zhenyu Wen, Tejal Shah, Graham Morgan, and Rajiv Ranjan.
\newblock Explainable {AI} ({XAI}): Core ideas, techniques, and solutions.
\newblock 55\penalty0 (9):\penalty0 194:1--194:33.
\newblock ISSN 0360-0300.
\newblock \doi{10.1145/3561048}.

\bibitem[Lundberg and Lee()]{Lundberg20174766}
Scott~M. Lundberg and Su-In Lee.
\newblock A unified approach to interpreting model predictions.
\newblock In \emph{Proceedings of the 31st International Conference on Neural Information Processing Systems}, {NIPS}'17, pages 4768--4777. Curran Associates Inc.
\newblock ISBN 978-1-5108-6096-4.

\bibitem[Ribeiro et~al.({\natexlab{a}})Ribeiro, Singh, and Guestrin]{ribeiro2016should}
Marco~Tulio Ribeiro, Sameer Singh, and Carlos Guestrin.
\newblock "why should i trust you?": Explaining the predictions of any classifier.
\newblock In \emph{Proceedings of the 22nd {ACM} {SIGKDD} International Conference on Knowledge Discovery and Data Mining}, {KDD} '16, pages 1135--1144. Association for Computing Machinery, {\natexlab{a}}.
\newblock ISBN 978-1-4503-4232-2.
\newblock \doi{10.1145/2939672.2939778}.

\bibitem[Friedman()]{friedman2001greedy}
Jerome~H. Friedman.
\newblock Greedy function approximation: A gradient boosting machine.
\newblock 29\penalty0 (5):\penalty0 1189--1232.
\newblock ISSN 0090-5364.
\newblock URL \url{https://www.jstor.org/stable/2699986}.
\newblock Publisher: Institute of Mathematical Statistics.

\bibitem[Ribeiro et~al.({\natexlab{b}})Ribeiro, Singh, and Guestrin]{ribeiro_anchors_2018}
Marco~Tulio Ribeiro, Sameer Singh, and Carlos Guestrin.
\newblock Anchors: High-precision model-agnostic explanations.
\newblock 32\penalty0 (1), {\natexlab{b}}.
\newblock ISSN 2374-3468.
\newblock \doi{10.1609/aaai.v32i1.11491}.
\newblock Number: 1.

\bibitem[Chipman et~al.()Chipman, George, and {McCulloch}]{chipman1998making}
H.~Chipman, Edward George, and Robert {McCulloch}.
\newblock Making sense of a forest of trees.
\newblock 29.

\bibitem[Haddouchi and Berrado()]{Haddouchi2019}
Maissae Haddouchi and Abdelaziz Berrado.
\newblock A survey of methods and tools used for interpreting random forest.
\newblock In \emph{2019 1st International Conference on Smart Systems and Data Science ({ICSSD})}, pages 1--6.
\newblock \doi{10.1109/ICSSD47982.2019.9002770}.

\bibitem[Aria et~al.()Aria, Cuccurullo, and Gnasso]{aria2021comparison}
Massimo Aria, Corrado Cuccurullo, and Agostino Gnasso.
\newblock A comparison among interpretative proposals for random forests.
\newblock 6:\penalty0 100094.
\newblock ISSN 2666-8270.
\newblock \doi{10.1016/j.mlwa.2021.100094}.

\bibitem[Mashayekhi and Gras()]{Mashayekhi2015223}
Morteza Mashayekhi and Robin Gras.
\newblock Rule extraction from random forest: the {RF}+{HC} methods.
\newblock In Denilson Barbosa and Evangelos Milios, editors, \emph{Advances in Artificial Intelligence}, Lecture Notes in Computer Science, pages 223--237. Springer International Publishing.
\newblock ISBN 978-3-319-18356-5.
\newblock \doi{10.1007/978-3-319-18356-5_20}.

\bibitem[Hara and Hayashi()]{hara2018making}
Satoshi Hara and Kohei Hayashi.
\newblock Making tree ensembles interpretable: A bayesian model selection approach.
\newblock In \emph{Proceedings of the Twenty-First International Conference on Artificial Intelligence and Statistics}, pages 77--85. {PMLR}.
\newblock {ISSN}: 2640-3498.

\bibitem[Zhao et~al.()Zhao, Wu, Lee, and Cui]{Zhao2019407}
Xun Zhao, Yanhong Wu, Dik~Lun Lee, and Weiwei Cui.
\newblock {iForest}: Interpreting random forests via visual analytics.
\newblock 25\penalty0 (1):\penalty0 407--416.
\newblock ISSN 1941-0506.
\newblock \doi{10.1109/TVCG.2018.2864475}.
\newblock Conference Name: {IEEE} Transactions on Visualization and Computer Graphics.

\bibitem[Hatwell et~al.()Hatwell, Gaber, and Azad]{Hatwell20205747}
Julian Hatwell, Mohamed~Medhat Gaber, and R.~Muhammad~Atif Azad.
\newblock {CHIRPS}: Explaining random forest classification.
\newblock 53\penalty0 (8):\penalty0 5747--5788.
\newblock ISSN 1573-7462.
\newblock \doi{10.1007/s10462-020-09833-6}.

\bibitem[Neto and Paulovich()]{neto2020explainable}
Mário~Popolin Neto and Fernando~V. Paulovich.
\newblock Explainable matrix - visualization for global and local interpretability of random forest classification ensembles.
\newblock 27\penalty0 (2):\penalty0 1427--1437.
\newblock ISSN 1077-2626.
\newblock \doi{10.1109/TVCG.2020.3030354}.
\newblock Publisher: {IEEE} Computer Society.

\bibitem[Dedja et~al.()Dedja, Nakano, Pliakos, and Vens]{Dedja20231}
Klest Dedja, Felipe~Kenji Nakano, Konstantinos Pliakos, and Celine Vens.
\newblock {BELLATREX}: Building explanations through a {LocaLly} {AccuraTe} rule {EXtractor}.
\newblock 11:\penalty0 41348 -- 41367.
\newblock ISSN 2169-3536.
\newblock \doi{10.1109/ACCESS.2023.3268866}.

\bibitem[Van~Assche and Blockeel(2007)]{van2007seeing}
Anneleen Van~Assche and Hendrik Blockeel.
\newblock Seeing the forest through the trees: Learning a comprehensible model from an ensemble.
\newblock In \emph{Machine Learning: ECML 2007: 18th European Conference on Machine Learning, Warsaw, Poland, September 17-21, 2007. Proceedings 18}, pages 418--429. Springer, 2007.

\bibitem[Zhou and Hooker(2016)]{zhou2016interpreting}
Yichen Zhou and Giles Hooker.
\newblock Interpreting models via single tree approximation, 2016.

\bibitem[Deng()]{deng2019interpreting}
Houtao Deng.
\newblock Interpreting tree ensembles with {inTrees}.
\newblock 7\penalty0 (4):\penalty0 277--287.
\newblock ISSN 2364-4168.
\newblock \doi{10.1007/s41060-018-0144-8}.

\bibitem[Gulowaty and Woźniak()]{gulowaty_extracting_2021}
Bogdan Gulowaty and Michał Woźniak.
\newblock Extracting interpretable decision tree ensemble from random forest.
\newblock In \emph{2021 International Joint Conference on Neural Networks ({IJCNN})}, pages 1--8.
\newblock \doi{10.1109/IJCNN52387.2021.9533601}.
\newblock {ISSN}: 2161-4407.

\bibitem[Nakahara et~al.()Nakahara, Jinguji, Sato, and Sasao]{Nakahara2017266}
Hiroki Nakahara, Akira Jinguji, Simpei Sato, and Tsutomu Sasao.
\newblock A random forest using a multi-valued decision diagram on an {FPGA}.
\newblock In \emph{2017 {IEEE} 47th International Symposium on Multiple-Valued Logic ({ISMVL})}, pages 266--271.
\newblock \doi{10.1109/ISMVL.2017.40}.
\newblock {ISSN}: 2378-2226.

\bibitem[Silva et~al.()Silva, Silva, Moreira, Nacif, and Ferreira]{silva_rdsf_2023}
Olavo A.~B. Silva, Alysson K.~C. Silva, Ícaro G.~S. Moreira, José A.~M. Nacif, and Ricardo~S. Ferreira.
\newblock {RDSF}: Everything at same place all at once - a random decision single forest.
\newblock In \emph{Anais do Simpósio Brasileiro de Engenharia de Sistemas Computacionais ({SBESC})}, pages 31--36. {SBC}.
\newblock {ISSN}: 2237-5430.

\bibitem[Murtovi et~al.()Murtovi, Bainczyk, Nolte, Schlüter, and Steffen]{murtovi_forest_2023}
Alnis Murtovi, Alexander Bainczyk, Gerrit Nolte, Maximilian Schlüter, and Bernhard Steffen.
\newblock Forest {GUMP}: a tool for verification and explanation.
\newblock 25\penalty0 (3):\penalty0 287--299.
\newblock ISSN 1433-2787.
\newblock \doi{10.1007/s10009-023-00702-5}.

\bibitem[Oliver(1993)]{oliver1992decision}
Jonathan~J. Oliver.
\newblock Decision graphs - an extension of decision trees.
\newblock 1993.

\bibitem[Ignatov and Ignatov()]{8372043}
Dmitry Ignatov and Andrey Ignatov.
\newblock Decision stream: Cultivating deep decision trees.
\newblock In \emph{2017 {IEEE} 29th International Conference on Tools with Artificial Intelligence ({ICTAI})}, pages 905--912.
\newblock \doi{10.1109/ICTAI.2017.00140}.
\newblock {ISSN}: 2375-0197.

\bibitem[Needham and Dowe()]{pmlr-vR3-needham01a}
Scott Needham and David~L. Dowe.
\newblock Message length as an effective ockham’s razor in decision tree induction.
\newblock In \emph{International Workshop on Artificial Intelligence and Statistics}, pages 216--223. {PMLR}.
\newblock {ISSN}: 2640-3498.

\bibitem[Tan and Dowe()]{Tan2002131}
Peter~J. Tan and David~L. Dowe.
\newblock {MML} inference of decision graphs with multi-way joins and dynamic attributes.
\newblock In Tamás (Tom)~Domonkos Gedeon and Lance Chun~Che Fung, editors, \emph{{AI} 2003: Advances in Artificial Intelligence}, Lecture Notes in Computer Science, pages 269--281. Springer.
\newblock ISBN 978-3-540-24581-0.
\newblock \doi{10.1007/978-3-540-24581-0_23}.

\bibitem[Florio et~al.()Florio, Martins, Schiffer, Serra, and Vidal]{Florio20237577}
Alexandre~M. Florio, Pedro Martins, Maximilian Schiffer, Thiago Serra, and Thibaut Vidal.
\newblock Optimal decision diagrams for classification.
\newblock 37\penalty0 (6):\penalty0 7577--7585.
\newblock ISSN 2374-3468.
\newblock \doi{10.1609/aaai.v37i6.25920}.
\newblock Number: 6.

\bibitem[Zhu and Shoaran()]{Zhu202113707}
Bingzhao Zhu and Mahsa Shoaran.
\newblock Tree in tree: from decision trees to decision graphs.
\newblock In \emph{Advances in Neural Information Processing Systems}, volume~34, pages 13707--13718. Curran Associates, Inc.

\bibitem[Wallace()]{wallace2005statistical}
Christopher~Stewart Wallace.
\newblock \emph{Statistical and Inductive Inference by Minimum Message Length}.
\newblock Information Science and Statistics. Springer-Verlag.
\newblock ISBN 978-0-387-23795-4.
\newblock \doi{10.1007/0-387-27656-4}.

\bibitem[Brandes()]{brandes_variants_2008}
Ulrik Brandes.
\newblock On variants of shortest-path betweenness centrality and their generic computation.
\newblock 30\penalty0 (2):\penalty0 136--145.
\newblock ISSN 0378-8733.
\newblock \doi{10.1016/j.socnet.2007.11.001}.

\bibitem[Mones et~al.()Mones, Vicsek, and Vicsek]{mones_hierarchy_2012}
Enys Mones, Lilla Vicsek, and Tamás Vicsek.
\newblock Hierarchy measure for complex networks.
\newblock 7\penalty0 (3):\penalty0 e33799.
\newblock ISSN 1932-6203.
\newblock \doi{10.1371/journal.pone.0033799}.
\newblock Publisher: Public Library of Science.

\bibitem[Radicchi et~al.()Radicchi, Castellano, Cecconi, Loreto, and Parisi]{radicchi_defining_2004}
Filippo Radicchi, Claudio Castellano, Federico Cecconi, Vittorio Loreto, and Domenico Parisi.
\newblock Defining and identifying communities in networks.
\newblock 101\penalty0 (9):\penalty0 2658--2663.
\newblock \doi{10.1073/pnas.0400054101}.
\newblock Publisher: Proceedings of the National Academy of Sciences.

\bibitem[Raghavan et~al.()Raghavan, Albert, and Kumara]{lpa_raghavan_near_2007}
Usha~Nandini Raghavan, Réka Albert, and Soundar Kumara.
\newblock Near linear time algorithm to detect community structures in large-scale networks.
\newblock 76\penalty0 (3):\penalty0 036106.
\newblock \doi{10.1103/PhysRevE.76.036106}.
\newblock Publisher: American Physical Society.

\bibitem[Scornet()]{scornet_trees_2021}
Erwan Scornet.
\newblock Trees, forests, and impurity-based variable importance.

\end{thebibliography}

\end{document}